\documentclass[10pt,twocolumn,letterpaper]{article}

\usepackage{cvpr}              

\usepackage[utf8]{inputenc}
\usepackage{dblfloatfix}
\usepackage{float}         
\usepackage{array}
\usepackage{tabularx}
\usepackage{booktabs}
\usepackage{multirow}
\usepackage{xcolor}
\usepackage{colortbl}
\usepackage{amssymb}
\usepackage{pifont}
\usepackage{comment}
\usepackage{tikz}
\usepackage{graphicx}
\usetikzlibrary{positioning, shapes.geometric, shapes.arrows, arrows.meta}
\definecolor{headerblue}{RGB}{65, 105, 225}
\definecolor{highlightyellow}{RGB}{255, 255, 153}
\usetikzlibrary{shapes.arrows, shapes.symbols, positioning, shadows, shadows.blur, decorations.pathmorphing, patterns, arrows.meta}
\usepackage[most]{tcolorbox}
\usepackage{geometry}

\newcommand{\best}[1]{\textbf{\color{blue!70!black}#1}}

\definecolor{cvprblue}{rgb}{0.21,0.49,0.74}
\usepackage[pagebackref,breaklinks,colorlinks,allcolors=cvprblue]{hyperref}

\def\paperID{17281} 
\def\confName{CVPR}
\def\confYear{2026}

\title{Narrative Aligned Long Form Video Question Answering}

\author{
Rahul Jain$^{1}$\thanks{Work done during internship at Amazon.}, \quad Keval Doshi$^{2}$, \quad Burak Uzkent$^{2}$, \quad Garin Kessler$^{2}$\\
$^1$Purdue University
\quad $^2$ Amazon \\
{\tt\small jain@348@purdue.edu}; \quad  
{\tt\small \{kcdos, burauzke, kesslerg\}@amazon.com}
}

\begin{document}
\maketitle
\begin{abstract}
Recent progress in MLLMs has led to a surge of benchmarks for long-video reasoning, yet most rely on localized cues and fail to capture narrative reasoning—the ability to track intentions, connect distant events, and reconstruct causal chains across an entire movie. We introduce NA-VQA, a benchmark built to evaluate deep temporal and narrative reasoning in long-form videos. NA-VQA includes 88 full movies and 4.4K open-ended QA pairs, each grounded in multiple evidence spans labeled as Short, Medium, or Far to assess long-range dependencies. By requiring generative, multi-scene answers, NA-VQA tests whether models can integrate dispersed narrative clues instead of relying on pattern matching. To overcome the limitations of existing models, we propose Video-NaRA, a narrative-centric framework that constructs event-level chains and stores them in a narrative memory for retrieval during reasoning. Across extensive evaluations, state-of-the-art MLLMs consistently fail on questions requiring Far evidence, highlighting the need for explicit narrative modeling. Video-NaRA improves long-range reasoning by up to 3\%, demonstrating its effectiveness in handling dispersed narrative structure. We will release our dataset NA-VQA fully upon publication.
\end{abstract}
\section{Introduction}
\label{sec:intro}

\begin{figure}
    \centering
    \includegraphics[width=\linewidth]{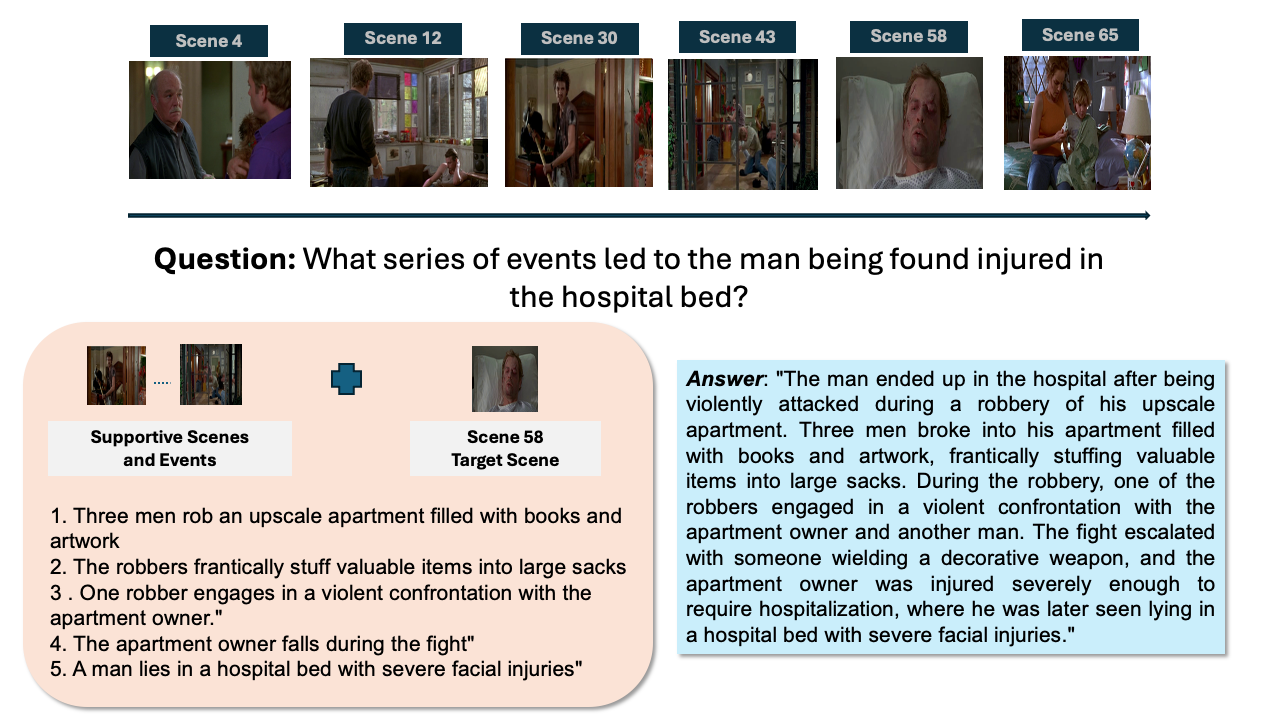}
    \caption{Illustration of \textbf{NA-VQA} sample. NA-VQA consists of a question from a full movie, the key evidence scenes collected from different parts of the timeline, and the final answer. This example shows how NA-VQA tests a model’s ability to connect scattered events and recover the full sequence of what happened.}
    \label{fig:enter-label}
\end{figure}

In recent years, multimodal large language models (MLLMs) have achieved significant progress in understanding long-form videos \cite{Song2023MovieChatFD, Li2023VideoChatCV, Bai2025Qwen25VLTR}. This has motivated the development of 
benchmarks to evaluate their reasoning abilities on long video content. Several recent works, including CinePile \cite{Rawal2024CinePileAL}, MH-VidQA \cite{Chen2024GroundedMV}, CG-Bench \cite{Chen2024CGBenchCQ}, and Rextime \cite{Chen2024ReXTimeAB}, have begun exploring temporal reasoning in videos.
 
Temporal reasoning in real-world long videos involves understanding complex, dynamic changes that unfold over time. Models must track events that are temporally distant, follow how intentions evolve, and reason about how actions lead to outcomes. More importantly, the information required to answer a question is rarely localized - it is often scattered across the timeline, with temporal relationships emerging gradually through narrative progression. While prior benchmarks such as CG-Bench \cite{Chen2024CGBenchCQ} and Rextime \cite{Chen2024ReXTimeAB} explore temporal understanding, they primarily focus on single isolated clues within short videos ($\leq 1$ hour), falling short of capturing the complexity of full-length narrative reasoning. Moreover, these benchmarks mainly use multiple-choice format, where models can exploit answer priors without fully comprehending or engaging with video content. However, what remains missing is a \emph{benchmark} that requires models to construct reasoning chains across multiple, temporally distant events - mirroring how humans piece together narratives from key moments distributed throughout a story.

\begin{figure*}
    \includegraphics[width=0.8\linewidth]{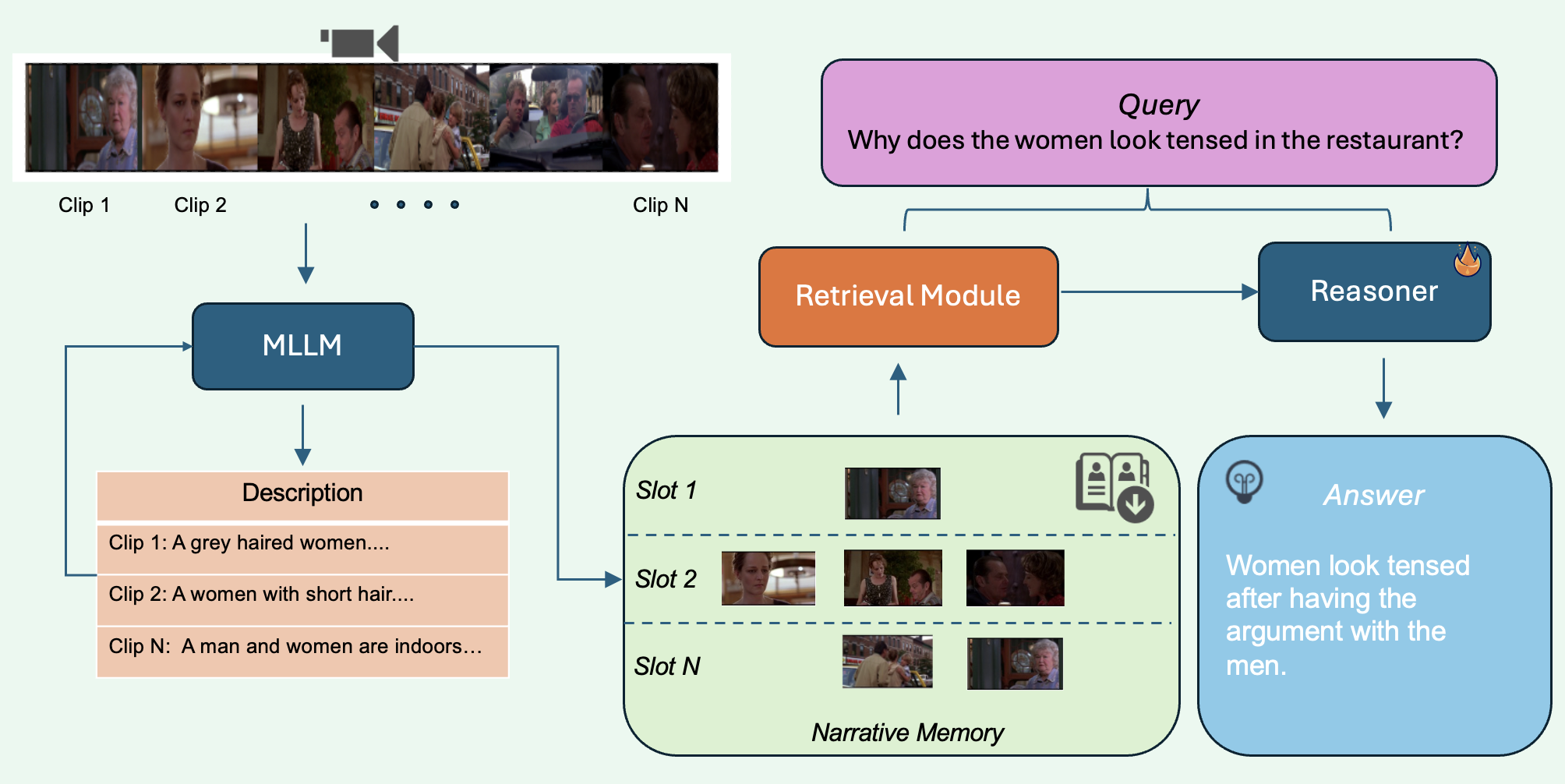}
    \centering
    \caption{\textbf{Video-NaRA pipeline}. We begin by generating a short description for every clip using Qwen-VL 2.5 (7B). Using both the clip and its description, the model groups clips into narrative slots and builds a structured narrative memory. Given a query, the retrieval module selects the most relevant clips, which are then passed to a fine-tuned reasoner (Qwen-VL 2.5 7B) to produce the final answer. }
    \label{fig:workflow}
\end{figure*}

To address these limitations, we introduce Narrative-Aligned Video Question Answering (NA-VQA), a comprehensive benchmark designed to evaluate the reasoning capabilities of MLLMs in long-form video settings. Our proposed dataset comprises of 88 complete movies spanning diverse genres and narrative styles, with question-answer pairs categorized into seven reasoning types: Causal, Narrative, Character-centric, Thematic, Goal-based, Social, and Hypothetical. Unlike prior work, each question in the dataset is explicitly grounded in multiple evidence spans distributed across the video timeline. We categorize these spans by temporal distance into \emph{Short, Medium}, and \emph{Far}, enabling fine-grained analysis of how models handle long-range dependencies. By requiring generative answers grounded in multi-scene evidence, NA-VQA ensures models demonstrate genuine temporal reasoning rather than exploiting surface-level patterns or answer priors. 

To establish baseline performance and understand current model limitations, we conduct extensive evaluation of state-of-the-art MLLMs on NA-VQA, including both short-context models and recent long-video architectures \cite{Chen2024CGBenchCQ, Chen2024ReXTimeAB}. Our analysis reveals systematic failures: models struggle particularly with questions requiring \emph{Far} temporal evidence, where relevant scenes are separated by extended durations. Qualitative analysis shows that models tend to fail to establish causal links between temporally distant events. These findings suggest that existing architectures, which process videos as uniform sequences of frames, lack explicit mechanisms for maintaining and reasoning over narrative structures that span entire movies.

Motivated by these limitations, and inspired by how humans construct narrative understanding—by identifying key events and building causal chains rather than processing every frame uniformly—we propose Video-NaRA (Narrative Reasoning and Alignment).  It constructs narrative chains from the video and store them in the memory. Finally, relevant narratives are retrieved for the reasoning model to answer the question. Additionally, We perform extensive benchmarking of the existing MLLM models with both short and long form video Models. From our experiments, NaRA demonstrates relative performance improvements of up to 3\% compared to baseline approaches. 

In summary, we make the following contributions:

 \begin{itemize}
 \item A novel taxonomy covering 7 reasoning types from temporal localization to complex causal chains spanning entire narratives in the long form video. 
 \item A memory-augmented architecture utilizing query-conditioned vision storage in memory and small reasoning LLMs for structured output generation.
\item We provide extensive benchmarking and ablations of state-of-the-art VideoQA models on our datasets.
\item We will release our NA-VQA dataset to the public upon publication.
 \end{itemize}

\begin{table*}[ht]
\centering
\scriptsize
\setlength{\tabcolsep}{4pt}
\renewcommand{\arraystretch}{1.2}
\begin{tabular}{l|r|r|c|c|ccccccc}
\toprule
\multirow{2}{*}{\textbf{Dataset}} &
\multirow{2}{*}{\textbf{\begin{tabular}[c]{@{}c@{}}QA \\ Samples \end{tabular}}} &
\multirow{2}{*}{\textbf{\begin{tabular}[c]{@{}c@{}}Avg. Duration \\ (sec) \end{tabular}}} &
\multirow{2}{*}{\textbf{\begin{tabular}[c]{@{}c@{}}Multi-hop\end{tabular}}} &
\multirow{2}{*}{\textbf{\begin{tabular}[c]{@{}c@{}}Evidence\\Local.\end{tabular}}} &
\multicolumn{7}{c}{\textbf{Reasoning Types}} \\
\cmidrule(lr){6-12}
& & & & &
\textbf{Goal.} &
\textbf{Caus.} &
\textbf{Hypo.} &
\textbf{Char.} &
\textbf{Narr.} &
\textbf{Theme} &
\textbf{Social}\\
\midrule
\rowcolor{gray!5}
MoVQA \cite{zhang2023movqa}& 21,953 & 992 & \checkmark & \checkmark & -- & \checkmark & \checkmark  & -- & -- & -- & --\\
RexTime \cite{Chen2024ReXTimeAB}& 2,143 & 141.1 & -- & \checkmark & --& \checkmark & -- & -- & -- & -- & --\\
MultiHop-EgoQA \cite{Chen2024GroundedMV} & -- & 180 & \checkmark & \checkmark & \checkmark & -- & -- & -- & -- & -- & --\\
\rowcolor{gray!5}
IntentQA \cite{Li2023IntentQACV} & 16,297 & unk. & -- & -- & \checkmark & \checkmark & -- & -- & -- & -- & -- \\
Video-MME\cite{Fu2024VideoMMETF} & 2,700 & 1,017.9 & -- & -- & \checkmark & -- & \checkmark & \checkmark & -- & -- & --\\
\rowcolor{gray!5}
MVBench \cite{Li2023MVBenchAC}& 4,000 & 16 & -- & -- & \checkmark & -- & \checkmark & \checkmark & -- & -- & --\\
Video-Bench  \cite{Ning2023VideoBenchAC}& 17,036 & 56 & -- & -- & \checkmark & -- & -- & -- & -- & -- & -- \\
\rowcolor{gray!5}
LVBench \cite{Wang2024LVBenchAE}& 1,549 & 4,101 & -- & -- & -- & -- & -- & \checkmark &  -- & -- & --\\
MLVU \cite{Zhou2024MLVUBM}& 2,000 & 930 & -- & -- & \checkmark & \checkmark & -- & \checkmark & \checkmark & -- & -- \\
\rowcolor{gray!5}
CinePile \cite{Rawal2024CinePileAL}& 303,828 & 160 & -- & -- & \checkmark & \checkmark & -- & \checkmark & \checkmark & \checkmark & --  \\
Neptune \cite{Nagrani2024NeptuneTL}& 150 & 2,405 & -- & -- & \checkmark & \checkmark & \checkmark & -- & \checkmark & -- & --\\
\rowcolor{gray!5}
CG-Bench \cite{Chen2024CGBenchCQ}& 12,129 & 1,624.4 & -- & \checkmark & \checkmark & -- & -- & \checkmark & -- & -- & --\\
VR-Bench \cite{Yu2025VRBenchAB}& 9,468 & 5,796 & \checkmark & \checkmark & \checkmark & \checkmark & \checkmark & \checkmark & \checkmark & -- & -- \\
MovieCORE \cite{Faure2025MovieCORECR} & 4,930 & 600 & \checkmark & \checkmark & \checkmark & \checkmark & -- & \checkmark & \checkmark & \checkmark & \checkmark \\
SeriesBench \cite{Zhang2025SeriesBenchAB} & 29,196 & 79 & -- & \checkmark & \checkmark & \checkmark & -- & \checkmark & \checkmark & \checkmark & -- \\
\midrule
\rowcolor{blue!10}
\textbf{NA-VQA} & \textbf{4442} & \textbf{2 hrs} & \cellcolor{green!20}\checkmark & \cellcolor{green!20}\checkmark & \cellcolor{green!20}\checkmark & \cellcolor{green!20}\checkmark & \cellcolor{green!20}\checkmark & \cellcolor{green!20}\checkmark & \cellcolor{green!20}\checkmark & \cellcolor{green!20}\checkmark &
\cellcolor{green!20}\checkmark\\
\bottomrule
\end{tabular}
\caption{\textbf{Comparison of video QA datasets.} Our dataset provides comprehensive coverage across all reasoning types and evaluation dimensions. \textit{Abbreviations:} \#QA=Number of QA pairs, Avg=Average video duration, Multi-hop=Multihop reasoning, Evidence Local.=Evidence Localization, Goal.=Goal-based, Caus.=Causal, Hypo.=Hypothetical, Char.=Character, Narr.=Narrative, Theme=Thematic, Social=Social reasoning \checkmark~indicates feature presence, -- indicates absence.}
\label{tab:dataset_comparison}
\end{table*}

\section{Related Work}
\paragraph{Long Form Video Reasoning Benchmark} Recently, there is a shift towards long form video understanding \cite{Liu2023MMBenchIY, Liu2024TempCompassDV} especially focusing on reasoning tasks \cite{Chen2024CGBenchCQ, Fu2024VideoMMETF, Yu2025VRBenchAB, Zhou2024MLVUAC, Rawal2024CinePileAL}. The key factor for accurate answering in these tasks is the model's ability to locate the correct supporting events with sufficient evidence. Benchmarks such as EgoSchema \cite{Mangalam2023EgoSchemaAD} and Video-MME \cite{Fu2024VideoMMETF} focus on fine grained temporal understanding testing how well models can understand the timing and order of events in videos. CG-Bench \cite{Chen2024CGBenchCQ} emphasizes evidence-grounded reasoning by requiring models to identify supporting moments. More recently, VR-Bench \cite{Yu2025VRBenchAB} introduces reasoning with relevant clues, but focuses mainly on surface-level or direct visual reasoning. CinePile \cite{Rawal2024CinePileAL} focuses on some complex reasoning types such as narratives themes etc but misses evidence grounding. Recently, MovieCORE~\cite{Faure2025MovieCORECR} connects temporal events for reasoning however the length of videos are 10 min. In contrast, we propose a novel benchmark that provides rich, grounded visual clues/evidences aimed at evaluating more challenging forms of reasoning, such as narrative, thematic, and causal inference which enables a deeper assessment of a model’s ability to reason beyond visible cues for long form video (2 hrs). Table~\ref{tab:dataset_comparison} shows detailed statistics of our dataset and the differences between ours and existing benchmarks.
\vspace{-5mm}
\paragraph{MLLMs on Long-Form Videos}
As long form video contains long context, MLLMs are increasingly being utilized to tackle long-form video comprehension tasks. Video-MLLMs \cite{Li2023VideoChatCV, Li2024VideoChatFlashHC} leverage compression techniques by reducing the number of tokens per frame to manage lengthy inputs. On the other hand \cite{Xue2024LongVILASL, Zhang2024LongCT} extends the context window in MLLM. Modern Video-MLLMs, such as InternVL2.5 \cite{Chen2024ExpandingPB} and Qwen2.5-VL \cite{Bai2025Qwen25VLTR}, advance video understanding by incorporating dynamic visual encoders that enables more efficient long form video processing. To boost reasoning capabilities, some MLLMs \cite{Chen2025ScalingRT, Peng2025LMMR1E3} have adopted advanced reinforcement learning frameworks, enabling more effective reasoning.
%
\vspace{-4mm}
\paragraph{Memory based architecture}
Memory based architectures retrieve relevant information given the query without passing the entire video, thus avoiding information loss due to token limitations. Common strategies involve storing raw forms of information like descriptions \cite{Long2025SeeingLR}, knowledge graphs \cite{Xu2025AMEMAM}, or latent features \cite{Liu2024MemLongMR, Song2023MovieChatFD}. Visual memory methods such as MovieChat \cite{Song2023MovieChatFD}, MA-LMM \cite{He2024MALMMML}, and Hermes \cite{Faure2024HERMESTL} compress frame features into memory modules for later retrieval. MovieChat \cite{Song2023MovieChatFD}, for example, uses cosine similarity to store only the most relevant frame features. Most of these methods focus on either episodic memory (storing observed events) or semantic memory (storing general knowledge), typically in a sequential format. In contrast, our method introduces narrative memory, where we store clips forming temporally coherent chains, capturing character goals, object interactions, and evolving story arcs, enabling structured retrieval that supports reasoning.

\vspace{-2mm}
\section{Constructing NA-VQA Dataset}
\label{sec:dataset}

\begin{figure*}
\begin{center}
\begin{tikzpicture}[
    scale=0.75,
    every node/.style={transform shape},
    process/.style={
        signal, 
        signal from=west, 
        signal to=east,
        minimum height=1.2cm,
        minimum width=2.8cm,
        align=center,
        font=\normalsize\bfseries,
        blur shadow={shadow blur steps=5, shadow xshift=2pt, shadow yshift=-2pt},
        draw=black!80,
        line width=1.2pt,
        inner sep=6pt,
        signal pointer angle=110
    },
    agent/.style={
        rectangle,
        rounded corners=6pt,
        minimum width=2cm,
        minimum height=0.8cm,
        draw=black!80,
        line width=1.5pt,
        fill=white,
        font=\normalsize\bfseries,
        blur shadow={shadow blur steps=5, shadow xshift=2pt, shadow yshift=-2pt},
        double=black!20,
        double distance=1pt
    },
    imagebox/.style={
        rectangle,
        rounded corners=8pt,
        inner sep=3pt,
        draw=black!70,
        line width=2pt,
        blur shadow={shadow blur steps=5, shadow xshift=3pt, shadow yshift=-3pt},
        fill=white
    },
    arrow/.style={
        ->, 
        line width=2pt, 
        draw=black!70,
        -{Stealth[length=3mm, width=2.5mm]}
    },
    backarrow/.style={
        ->, 
        line width=1.5pt, 
        draw=black!50, 
        dashed,
        dash pattern=on 4pt off 3pt,
        -{Stealth[length=2.5mm, width=2mm]}
    }
]

\node[process, fill=orange!80, text=white] (p1) at (0,0) {Raw Data\\Collection};

\node[process, fill=orange!50, anchor=west] (p2) at (p1.east) {Event Extraction\\from raw description};

\node[process, fill=red!30, anchor=west] (p3) at (p2.east) {Initial VQA};

\node[process, fill=blue!25, anchor=west] (p4) at (p3.east) {Deeper Analysis for\\improving VQA};

\node[process, fill=teal!35, anchor=west] (p5) at (p4.east) {Final Action and\\Improving VQA};

\node[process, fill=green!30, anchor=west] (p6) at (p5.east) {Final VQA};

\node[imagebox, above=0.9cm of p2] (imgbox1) {\includegraphics[width=1.5cm, height=1.5cm]{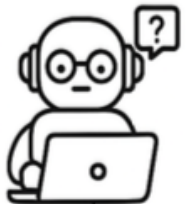}};
\node[agent, above=0.08cm of imgbox1, fill=orange!10] (label1) {Event Extractor};

\node[imagebox, above=0.9cm of p3] (imgbox2) {\includegraphics[width=1.5cm, height=1.5cm]{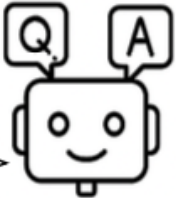}};
\node[agent, above=0.08cm of imgbox2, fill=red!10] (label2) {VQA Expert};

\node[imagebox, above=0.9cm of p4] (imgbox3) {\includegraphics[width=1.5cm, height=1.5cm]{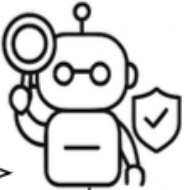}};
\node[agent, above=0.08cm of imgbox3, fill=blue!10] (label3) {VQA Validator};

\node[imagebox, above=0.9cm of p5] (imgbox4) {\includegraphics[width=1.5cm, height=1.5cm]{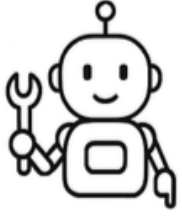}};
\node[agent, above=0.08cm of imgbox4, fill=teal!10] (label4) {VQA Refiner};

\draw[arrow, draw=orange!60] (imgbox1) -- (p2);
\draw[arrow, draw=red!60] (imgbox2) -- (p3);
\draw[arrow, draw=blue!60] (imgbox3) -- (p4);
\draw[arrow, draw=teal!60] (imgbox4) -- (p5);

\draw[backarrow] (p1.north) to[out=60, in=180] (imgbox1.west);

\draw[backarrow] (p2.north east) to[out=20, in=200] (imgbox2.west);

\draw[backarrow] (p3.north east) to[out=20, in=200] (imgbox3.west);

\draw[backarrow] (p4.north east) to[out=20, in=200] (imgbox4.west);

\end{tikzpicture}
\end{center}
 \caption{\textbf{NA-VQA} dataset creation pipeline. We start with raw movie data, extract event-level descriptions, generate an initial set of QA pairs using LLM, validate them through deeper analysis, and finally refine the QA outputs using an LLM-based refiner. This multi-stage process ensures high-quality, narrative-grounded VQA annotations for long-form movie videos.}
 \label{fig:stat}
\end{figure*}
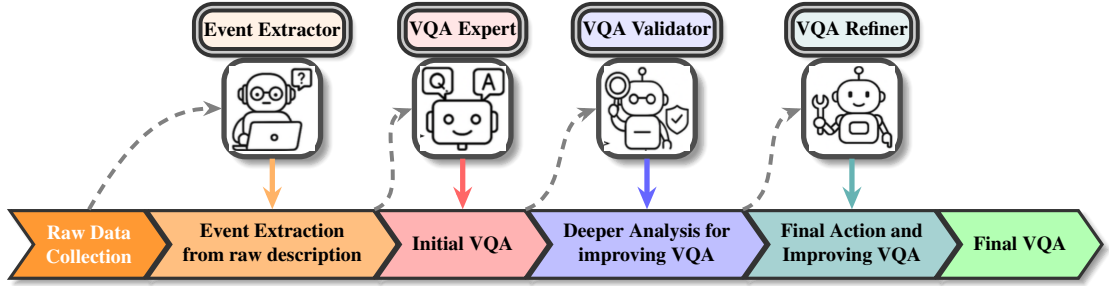

To construct NA-VQA, we first collect raw video–caption pairs from the LSMDC dataset~\cite{Rohrbach2016MovieD}. From each full-length movie, we extract key events guided by our reasoning taxonomy to serve as candidates for question–answer generation. To ensure dataset quality and grounding, we employ an LLM-based automatic pipeline consisting of \emph{validation} and \emph{refinement} stages. Finally, we present detailed dataset statistics and distributions across reasoning categories.
\subsection{Raw video collection and related data.} 
To study reasoning in long-form videos, we collected 88 full-length movies from the LSMDC dataset~\cite{Rohrbach2016MovieD}. Movies provide a rich testbed for video reasoning as they contain complex narrative structures, causal dependencies, and long-range temporal dynamics that span across scenes through the whole video. We selected movies from LSMDC~\cite{Rohrbach2016MovieD} as it offers manually annotated fine-grained shot segmentation that we use as the basis for building structured event representations. Using LSMDC brings several advantages. (1) it includes diverse genres and storytelling styles, which helps ensure coverage of different reasoning patterns, (2) MAD dataset provide aligned audio descriptions, dialogues, and subtitles, which we leverage for multimodal analysis. (3) prior efforts such as MovieBench~\cite{Wu2024MovieBenchAH} have already annotated hierarchical structures like shots and scenes for these movies, reducing the need for low-level manual effort and allowing us to focus on higher-level reasoning. These existing resources allow us to build a scalable and semi-automated pipeline while maintaining quality.

\subsection{Event Extraction and Structuring}

To extract visual events in the video, we first extract detailed shot level description using Claude Sonnet 4.  Then we combine this shot level detailed visual description with LSMDC-provided shot annotations, which include \emph{background details, narrative cues}, and \emph{visual context}. The model-generated visual description and original LSMDC dataset-provided descriptions are combined to form a comprehensive textual representation of each shot. This shot level detailed description provide information that includes emotions, background settings, style elements, character information and camera motion. These details are important for visual reasoning. We then aggregate shot level descriptions to construct scene-level summaries as shots are roughly 2-3 seconds and does not contain complete information. These scene level descriptions are then used to extract events. 
To capture the narrative flow of the scene, we segment each scene into a sequence of discrete events. Each event represents a \emph{meaningful character, action, decision}, or \emph{discovery} that contributes to the plot. We prompt an LLM, e.g. Claude Sonnet 4, with the scene-level summary to extract 1–5 key events wherever possible. We use the following format for consistency: \emph{[Who] does [What] to [Whom/Why/Where]}. This structure helps ground each event in terms of agents, actions, and context, making it suitable for downstream reasoning. For example, from a summary like \emph{Alice enters the room and confronts Bob about the missing file}, the model might extract the following: \emph{Alice enters the room and Alice confronts Bob about the missing file}. Each extracted event is temporally grounded to the scene and serves as an intermediate representation for later stages including complex question answer pair generation.
\begin{figure*}[htbp]
    \centering
    \scalebox{0.82}{%
    \begin{minipage}{\textwidth}
        \centering
        \begin{subfigure}[b]{0.33\textwidth}
            \centering
            \includegraphics[width=\textwidth]{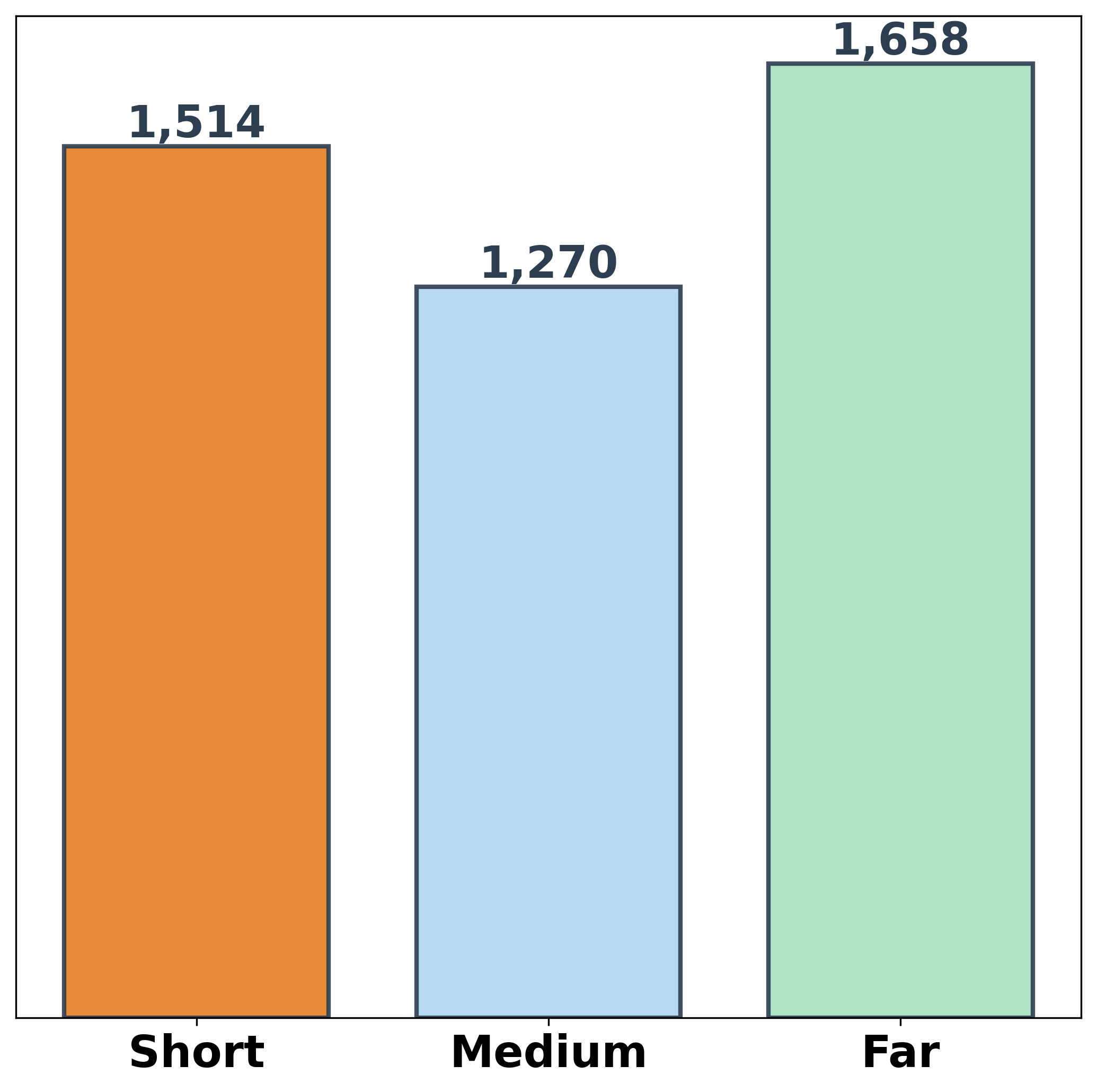}
            \caption{Scene Distance Distribution}
            \label{fig:scene_distance}
        \end{subfigure}
        \hfill
        \begin{subfigure}[b]{0.33\textwidth}
            \centering
            \includegraphics[width=\textwidth]{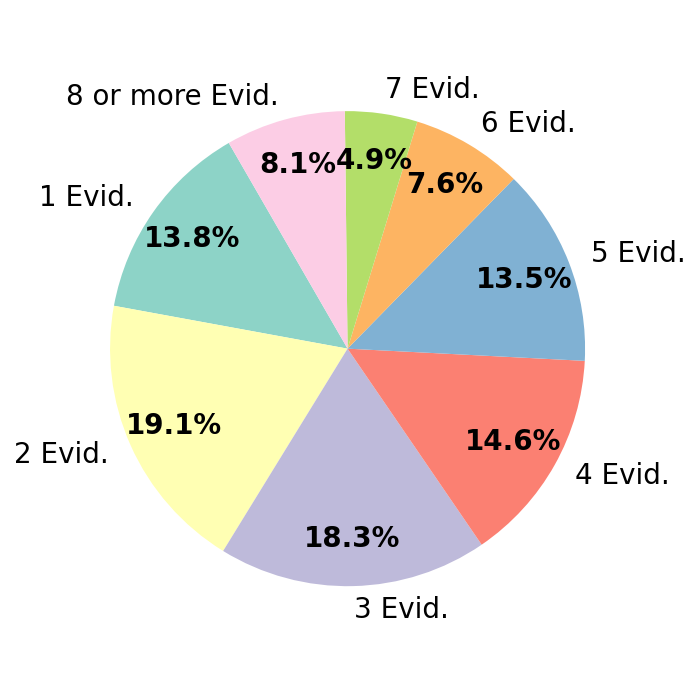}
            \caption{Evidence Scenes Distribution}
            \label{fig:evidence_scenes}
        \end{subfigure}
        \hfill
        \begin{subfigure}[b]{0.33\textwidth}
            \centering
            \includegraphics[width=\textwidth]{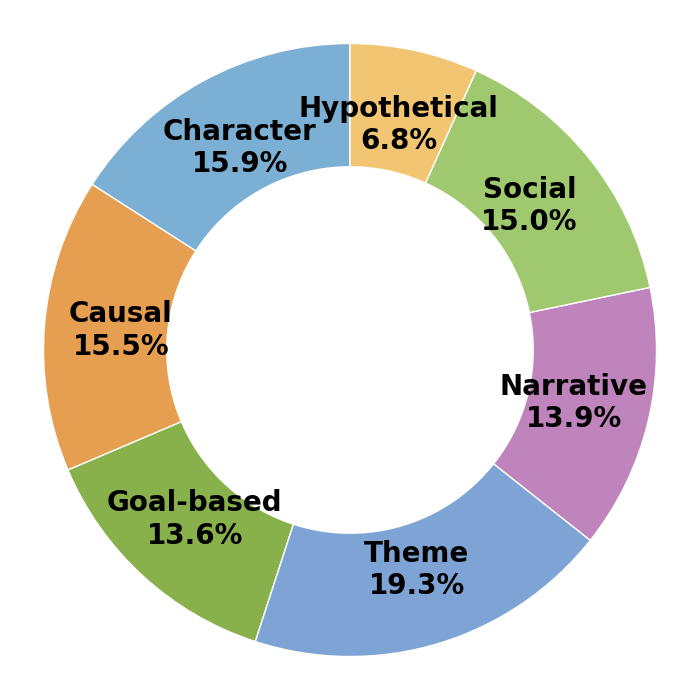}
            \caption{Reasoning Type Distribution}
            \label{fig:reasoning_type}
        \end{subfigure}
    \end{minipage}%
    }
    \caption{Comprehensive question analysis showing: (a) distribution by scene distance categories (short, medium, far), (b) distribution by number of evidence scenes required per question, and (c) distribution by reasoning type (Narrative, causal, theme etc.).}
    \label{fig:three_figures}
\end{figure*}


\subsection{Question and Answer Generation} 
Once the events are extracted, we prompt a language model to generate question-answer (QA) pairs along with supporting evidence. The goal is to create questions that require reasoning over multiple events rather than simple fact recall. We specifically instruct the LLM (Claude Sonnet 4) to generate multi-hop reasoning questions grounded in the extracted events and their temporal context. To structure this better, we organize QA generation along two key dimensions: (1) event distance (contextual span) and (2) reasoning type. Event distances categorizes into \emph{short}, \emph{medium} and \emph{far distance}. On the other hand, reasoning types are categorized into 7: \emph{Causal}, \emph{Narrative} \emph{Character}, \emph{Theme},\emph{Goal-Based}, \emph{Social}, and \emph{Hypothetical}. As the movies contain many reasoning aspects, these seven categories ensures that question and answer are generated correctly and adhere to reasoning categories. See details in the appendix.

\subsection{Automatic Data filtering} 
While the above process generates diverse and contextually rich question–answer pairs, several issues persist: (1) answers may rely on world knowledge rather than video evidence, (2) questions can be biased toward specific scenes or characters, (3) some pairs lack direct grounding in the visual content, and (4) evidence often includes non-visual cues such as character names.
To address these issues and improve the reliability of our dataset, we introduce an automated MLLM-based filtering pipeline consisting of two modules: (1) QA Validator and (2) QA Refiner. We use Claude Sonnet 4 as the LLM.  
\vspace{-5mm}
\paragraph{QA Validator}
The Validator ensures that each question–answer pair is truly grounded in the visual content rather than relying on textual or world knowledge. The model evaluates grounding based on three key principles:
(1) Character-Agnostic: names or textual identifiers are removed since they are not visual cues.
(2) Visual-Content Identifiability: the answer must be derivable from visible evidence, not inferred from background knowledge.
(3) Video-Grounded Framing: the phrasing of both question and answer should match what is actually observable in the scenes from the video.
Beyond grounding, the Validator focuses on the quality of evidence, since reasoning quality directly depends on how well the evidence supports the question–answer pair. We use three evidence-level criteria: \emph{\textbf{Completeness}}: The evidence must include all necessary visual segments that lead to the answer. Missing intermediate steps or transitions can make reasoning incomplete or disjointed, especially in long-form videos where dependencies unfold gradually.
\emph{\textbf{Minimality}}: The evidence should only include what is essential. Extra or unrelated events in the evidence introduce noise. 
\emph{\textbf{Faithfulness}}: The evidence must directly correspond to what the model observes and helps in answering the question, without introducing inferred or fabricated descriptions. 
Together, these principles enforce a reasoning chain of evidences that is tight, visual, and interpretable. For long videos, where temporal chains span across distant scenes, completeness, minimality, and faithfulness are crucial for ensuring that reasoning reflects true visual understanding rather than superficial correlations.
\vspace{-4mm}
\paragraph{QA Refiner}
The Refiner module improves the quality of question–answer pairs based on the Validator’s feedback. If the Validator assigns very low scores, the Refiner discards those pairs completely. For the remaining ones, it uses the Validator’s explanations and scores to make targeted edits. It modifies unclear or biased questions, removes any non-visual elements such as character names, and adjusts the answers to better match the verified evidence. Further it also improves evidence reliability. By applying these corrections, the Refiner strengthens the factual grounding and visual alignment of each QA pair, ensuring that the final dataset is both consistent and reliable.

\subsection{Dataset Statistics}

We present detailed statistics of our dataset to provide a clearer overview, including video meta-information, Question–Answer–Evidence (QAE) triplets, and qualitative breakdowns as illustrated in \autoref{fig:stat}. The NA-VQA dataset comprises 88 full-length movies from the LSMDC collection, each ranging from 80 to 200 minutes in duration, with the distribution shown in the figure. In total, the dataset contains 4,442 question–answer pairs supported by 17,327 evidence segments that span across entire movies. On average, each evidence set covers 64\%  of the scenes present in the video, indicating broad and sufficient coverage of the full narrative. The average number of evidences per question is 3.90. The dataset is uniformly distributed across all seven reasoning categories, ensuring balanced representation. We divide the 88 videos into a 2:1 train–test split, resulting in 59 movies for training and 29 movies for testing, corresponding to 2,967 and 1475 QA pairs in total.

\vspace{-2mm}
\section{Proposed Method (Video-NaRA)}
Humans naturally understand long videos by linking events through characters, objects, and context across time. Inspired by this, We propose a memory-based architecture that does not just store frames or captions in sequence. Instead, we build a narrative-driven structure that organizes the video based on high-level event chains. Our method first extracts key semantic information like which characters are involved, what their motivations are, and what intentions are driving the scene. Using this, we build narrative chains that connect related frames and events, grounded in shared characters, goals, and context. These chains are stored in memory and are retrieved during question answering. We retrieve only the relevant narrative chains and reason over them to find the answer. This lets us answer more complex “why” and “how” type questions by leveraging the structure of the story, not just the surface-level content.
\subsection{Narrative Memory Construction}
Narrative construction is critical for understanding long-form videos similar to humans. It involves connecting relevant sequences based on plot, character, objects, and context. To effectively construct such chains of temporal sequences, we leverage an MLLM (Qwen 2.5 VL 7B), which receives the input of raw frames and the descriptions of videos. The MLLM analyzes each clip sequentially to assign it to a memory, i.e., a narrative slot.

Given a video clip sequence $\mathcal{V} = {v_1, v_2, \dots, v_T}$ and corresponding textual descriptions $\mathcal{D} = {d_1, d_2, \dots, d_T}$, the MLLM processes each pair $(v_t, d_t)$ and assigns it to a narrative memory slot $s_i \in S$ based on temporal coherence and semantic similarity:
\begin{equation}
\text{MLLM}:(v_t, d_t) \rightarrow s_i, \qquad s_i \in S.
\end{equation}

This yields a narrative memory bank $\mathcal{M}$ formed by the union of all slots:
\begin{equation}
\mathcal{M} = \bigcup_{i=1}^{N} s_i
= \bigcup_{i=1}^{N} \left\{ (v_{t_j}, d_{t_j}) \right\}_{j=1}^{|s_i|}
\end{equation}

where $|s_i|$ denotes the number of clips assigned to slot $s_i$, and $N$ is the total number of slots. Each slot $s_i$ groups clips that follow a coherent storyline, capturing how characters, goals, and events evolve over time. This structure enables the model to retrieve evidence not as isolated frames but as narrative chains aligned with the underlying plot.

\subsection{Extracting Visual Features}
Existing video–text models often struggle with consistent clip-level alignment. To avoid noisy clip embeddings, we extract frame-level visual features first and then aggregate them to form a stable clip representation. For each frame $f_i$ in a clip $v_t$, we compute a dense visual embedding using a pretrained encoder $\phi_v$ (e.g., CLIP \cite{Radford2021LearningTV}):
\begin{equation}
\phi_v(f_i) \in \mathbb{R}^d.
\end{equation}

A clip $v_t$ is then represented by a set of features
$\mathcal{F}_{t} = \{\phi_v(f_1), \phi_v(f_2), \dots, \dots, \phi_v(f_{k})\}$ where $k$ is the number of sampled frames from clip $v_t$. These clip embeddings are then stored in their corresponding narrative memory slots $s_i \in \mathcal{M}$, where each slot groups clips that share semantic and temporal continuity and are used for retrieval task.

\subsection{Retrieval Module}
Given a query $q$, we first extract its feature representation $f_q = \phi_t(q)$ using the text encoder of the model employed for visual feature extraction (CLIP \cite{Radford2021LearningTV}). Our goal is to find the parts of the video that are most relevant to this query. Instead of searching through all clips independently, we take advantage of the narrative memory bank $\mathcal{M}$, which groups clips $\mathcal{V}$ into semantically coherent narrative slots $\mathcal{S}$. For each clip $v_t$, we compute a clip–query relevance score by measuring the cosine similarity between the query embedding and all frame embeddings in the clip, and then averaging the frame-level similarities:
\begin{equation}
z_t = \text{sim}(f_q, \mathbf{c}_t)
    = \mathbb{E}_{j \in v_t}\left[ \cos\left(f_q, f_j\right) \right].
\end{equation}

We then propagate these clip scores to the narrative level. Each slot
aggregates the relevance of its constituent clips, enabling the retrieval
module to reason over broader storyline structure. Each slot is assigned a narrative importance score $S_i$ using a
max-boosted weighted mean of the clip scores:
\begin{equation}
S_i = (1-\alpha)\,\text{mean}_{t \in s_i}(z_t)
    + \alpha\,\max_{t \in s_i}(z_t).
\end{equation}

To preserve fine-grained precision while exploiting narrative context,
each clip’s final relevance score is adjusted using its slot score:

\begin{equation}
r_t = z_t + \lambda S_i, \qquad v_t \in s_i.
\end{equation}

Finally, we rank all clips by $r_t$ and retrieve the top-$k$ clips as the
final evidence set. This narrative-aware scoring prefers clips that are
individually relevant while also being embedded within highly relevant
narrative chains, yielding temporally coherent evidence for downstream
reasoning.

\section{Experiments}
\label{sec:experiments}

\begin{table*}[ht]
\centering
\caption{Performance comparison across different distance ranges on NA-VQA benchmark. Best results excluding proprietary models are highlighted. You can find further results in Appendix.
}
\label{tab:performance}
\resizebox{\textwidth}{!}{
\setlength{\tabcolsep}{3pt}
\small
\begin{tabular}{@{}l|c|c|cccc|cccc|cccc|c@{}}
\toprule
\multirow{2.5}{*}{\textbf{Model}} &
\multirow{2.5}{*}{\textbf{\begin{tabular}{@{}c@{}}Use \\ Descr.\end{tabular}}} &
\multirow{2.5}{*}{\textbf{Frames}} & \multicolumn{4}{c|}{\textbf{Short}} & \multicolumn{4}{c|}{\textbf{Medium}} & \multicolumn{4}{c|}{\textbf{Far}} & \multirow{2.5}{*}{\textbf{Avg.}} \\
\cmidrule(lr){4-7} \cmidrule(lr){8-11} \cmidrule(lr){12-15} 
& & & Comp. & Depth & Evid. & Reas. & Comp. & Depth & Evid. & Reas. & Comp. & Depth & Evid. & Reas. & \\
\midrule
\rowcolor{gray!15}
\multicolumn{16}{c}{\textsc{\textbf{Proprietary Models}}} \\
\midrule
Claude Sonnet 4.5 &\checkmark & &  54.29 & 59.10 & 54.01 & 51.82 & 63.64 & 66.95 & 63.73 & 62.79 & 67.64 & 70.61 & 66.95 & 64.24 & 62.15 \\
GPT-120B OSS &\checkmark & & 45.78 & 50.14 & 43.87 & 43.71 & 49.96 & 53.94 & 46.91 & 48.84 & 51.89 & 57.18 & 50.31 & 50.04 & 49.38 \\
Qwen-235B & \checkmark& & 43.30 & 48.31 & 44.33 & 42.90 & 51.29 & 58.11 & 51.37 & 51.48 & 54.40 & 61.51 & 53.98 & 54.32 & 51.28 \\
\midrule
\rowcolor{gray!15}
\multicolumn{16}{c}{\textsc{\textbf{ZERO-SHOT VLMs}}} \\
\midrule
LLaVA-OneVision (7B) & & 32 & 28.68 & 31.94 & 23.39 & 24.32 & 30.04 & 33.22 & 24.64 & 25.55 & 31.43 & 35.17 & 26.22 & 27.38 & 28.50 \\
VideoLLaVA (7B) & & 8 & 18.62 & 21.14 & 17.23 & 14.18 & 16.57 & 19.61 & 15.67 & 12.45 & 15.79 & 17.84 & 14.90 & 11.55 & 16.30 \\
Video-LLaMA3 (7B) & & 128 &33.08 & 36.38 & 29.21 & 30.10 & 33.00 & 36.35 & 28.58 & 30.34 & 34.94 & 38.26 & 30.58 & 33.17 & 32.83 \\
VideoChatGPT (7B) & & 100 &27.82	&31.93&	22.93&	21.34&		28.76&	31.97&	23.82&	22.27&		27.66&	31.47&	23.86&	21.47& 26.28 \\
InternVL2 (7B) & &32 &28.39 & 29.98 & 25.83 & 25.21 & 29.96 & 31.29 & 26.82 & 26.82 & 29.65 & 32.36 & 28.03 & 26.22 & 28.38 \\
InternVL2.5 (7B) & &64 & 32.22 & 34.66 & 32.06 & 29.65 & 32.19 & 34.46 & 31.85 & 29.44 & 31.20 & 34.98 & 31.78 & 29.00 & 31.96 \\
Qwen2.5-VL (7B) & & 128& 32.54 & 35.83 & 29.00 & 29.65 & 33.60 & 37.33 & 30.34 & 31.71 & 34.58 & 39.03 & 32.16 & 32.37 & 33.18 \\
Qwen2.5 Instruct (7B) & \checkmark& & 25.99	&29.90	&25.54	&22.00	&	28.50	&31.93&	27.30&	25.58	&	25.52	&30.81	&25.83&	22.01 & 26.74 \\
\rowcolor{yellow!20}
\textbf{Video-NaRA} &\checkmark & 128
&34.26	&36.78	&32.38	&31.39 &35.92	&38.33&	33.69	&33.36 &35.10	&39.38	&34.58	&33.35 & 34.88 \\
\midrule
\rowcolor{gray!15}
\multicolumn{16}{c}{\textsc{\textbf{ZERO-SHOT Long-Form VLMs}}} \\
\midrule
MovieChat (7B) & & 2048 & 31.52 & 35.56 & 24.84 & 29.04 & 30.85 & 34.24 & 24.54 & 29.22 & 31.19 & 35.32 & 25.01 & 28.45 & 29.98 \\
VideoChat-Flash (7B) & & 3000 & 30.34 & 33.27 & 24.31 & 26.35 & 32.87 & 35.15 & 26.26 & 29.44 & 32.93 & 36.29 & 26.83 & 29.63 & 30.31 \\
\midrule
\rowcolor{gray!15}
\multicolumn{16}{c}{\textsc{\textbf{Fine-tuned Models}}} \\
\midrule
Qwen2.5-VL & & 128 & 32.63 & 36.21 & 29.74 & 30.79 & 35.02 & 38.41 & 31.55 & 33.49 & 34.55 & 38.80 & 31.88 & 32.38 & 33.79 \\
\rowcolor{yellow!20}
\textbf{Video-NaRA} &\checkmark & 128  &35.11	&38.37	&33.40	&32.59 & 36.18	&38.97&	34.98	&34.62 &35.98	&39.88	&35.10&	33.67 & 35.74 \\
\bottomrule
\end{tabular}}
\end{table*}

\subsection{Experimental Setups}

We conduct comprehensive experiments to evaluate Video-NaRA on its two core capabilities: (1) retrieving the correct evidence from long videos, and (2) reasoning over this evidence to answer open-ended questions. In addition, we compare Video-NaRA and benchmark a broad spectrum of Vision-Language Models (VLMs) and Large Language Models (LLMs)—covering both open-source and proprietary models to understand how current models perform on our proposed NA-VQA dataset.
\vspace{-4mm}
\paragraph{Baselines} For LLMs based evaluation we choose Claude Sonnet 4.5 \cite{anthropic2025claudeSonnet}, Qwen3-235B-A22B-Instruct \cite{qwen3technicalreport}, GPT-OSS-120B \cite{openai2025gptoss120bgptoss20bmodel} and Qwen 2.5 Instruct 7B \cite{Bai2025Qwen25VLTR}. To provide video information for these models, we segment each video into 32-second clips at 1 fps. These clips are passed through Qwen2.5-VL-72B to generate rich visual descriptions. We prompt the model with: \emph{Provide a detailed visual summary including key scenes, objects, people, and their interactions.}
The full visual description, along with the question, is then provided to the LLM for answer generation. We evaluate a comprehensive set of short-form and long-form VLMs models such as MovieChat~\cite{Song2023MovieChatFD} and VideoChatFlash~\cite{Li2023VideoChatCV}. Each model receives the raw video and the question as input and produces an answer directly. For both LLM and VLM evaluations, we adopt chain-of-thought prompting so models explicitly retrieve evidence first and then reason over it.
\vspace{-9mm}
\paragraph{Metrics} As NA-VQA contains open-ended questions, we adopt an LLM-as-judge evaluation protocol (claude Sonnet 4). For open-ended responses, following prior work \cite{Yu2025VRBenchAB, Faure2025MovieCORECR}, we use a judge model to assess answer quality across four dimensions: comprehensiveness, depth, evidence grounding, and reasoning quality. \emph{Depth Metric} evaluates models ability to connect events which reflect deeper understanding of underlying narrative. \emph{Comprehensive Metric} evaluate models ability to capture the main ideas and relevant context.  \emph{Evidence Metric} evaluates models ability to retrieve the evidences.  \emph{Reasoning Metric} evaluates models ability to infer relationship and logically consistent explanations. We use Claude Sonnet-4 as the evaluation judge.
\vspace{-4mm}
\paragraph{Implementation Details}
Videos are sampled at 1 fps and segmented into 32-second clips. Narrative memory is constructed offline using Qwen2.5-VL-7B. For final QA reasoning, we use Qwen2.5-VL-7B as the base model and finetune it with chain-of-thought on our training set focusing on evidence localization and answer generation. We fine-tune the model using 128 frames per sample with a batch size of 1 for one epoch. Video-NaRA selects the top 20 clips, which are uniformly sampled to 128 frames in total. 

\begin{table*}[]
\centering
\caption{Ablation results highlighting the contribution of each module in Video-NaRA Best results are in \textbf{bold}.}
\label{tab:ablation}
\resizebox{0.8\textwidth}{!}{
\setlength{\tabcolsep}{5pt}
\small
\begin{tabular}{l|cccc|cccc|cccc|}
\toprule
\multirow{2}{*}{\textbf{Model}} & \multicolumn{4}{c|}{\textbf{Short}} & \multicolumn{4}{c|}{\textbf{Medium}} & \multicolumn{4}{c|}{\textbf{Far}} \\
\cmidrule(lr){2-5} \cmidrule(lr){6-9} \cmidrule(lr){10-13} 
&Comp. & Depth & Evid. &Reas. & Comp. & Depth & Evid. &Reas.  & Comp. & Depth. & Evid. &Reas.\\
\midrule

{\color{blue}\ding{52}} \textbf{Full Model}& \textbf{35.11}	&\textbf{38.37}	&\textbf{33.40}	&\textbf{32.59} & \textbf{36.18}	&\textbf{38.97}&	\textbf{34.98}	&\textbf{34.62} &\textbf{35.98}	&\textbf{39.88}	&\textbf{35.10}&	\textbf{33.67}  \\
\addlinespace[3pt]
{\color{red}\ding{56}} w/o Narrative & 34.26&	36.78&	31.77&	31.73 & 36.35&	39.36	&34.08&	33.78& 35.16	&39.38&	32.70	&32.83  \\

\addlinespace[3pt]
{\color{red}\ding{56}} w/o CLIP & 34.26&	37.56	&31.85	&32.42& 35.54&	38.50	&32.92	&34.46 & 34.90&	39.46&	33.21&	32.24 \\

\addlinespace[3pt]
{\color{red}\ding{56}} w/o Both & 32.63 & 36.21 & 29.74 & 30.79 & 35.02 & 38.41 & 31.55 & 33.49 & 34.55 & 38.80 & 31.88 & 32.38  \\

\bottomrule
\end{tabular}}
\end{table*}
\vspace{-1mm}
\subsection{Results}
\subsubsection{Reasoning Evaluation — Multi-Step Processing}
We first evaluate how well different LLMs reason over multi-step descriptions. As shown in \autoref{tab:performance}, the proprietary model Claude Sonnet 4.5 achieves the highest performance across all evidence distances—Short, Medium, and Far—with an overall average score of 62\%. Among open-source models, GPT-OSS and Qwen3-235B-Instruct obtain average scores of 49.38 and 51.27, respectively. In contrast, Qwen2.5-VL-7B-Instruct performs the worst across all settings. Overall, Claude consistently delivers the strongest results across every metric, showing that it maintains tighter semantic alignment with ground-truth answers and retrieves supporting evidence more reliably than other alternatives.

\subsubsection{VLM Evaluation — Single-Step Processing}
We next evaluate Vision-Language Models (VLMs) on direct visual input. In the zero-shot setting, \autoref{tab:performance} shows that Qwen2.5-VL (7B) performs competitively among the baselines. Notably, InternVL2.5 (7B) achieves a comparable score using only 32 frames and 256 tokens, demonstrating stronger input efficiency. Long-form models such as MovieChat and VideoChat-Flash do not show performance gains even when scaled to 2048 frames, indicating that increasing frame count or reducing token budgets does not automatically translate to better reasoning. Video-NaRA achieves the best performance in the zero-shot setting. Fine-tuning on NA-VQA further improves the results for Qwen2.5-VL (7B), with the NA-VQA–finetuned model achieving the strongest overall performance.


\begin{table}[h]
\centering
\caption{Comparison across different reasoning categories. Numbers with~\best{color} show the best results.}
\label{tab:reasoning}
\scriptsize
\setlength{\tabcolsep}{3pt}
\renewcommand{\arraystretch}{1.3}
\begin{tabular}{l|ccccccc}
\toprule
\textbf{Model} & \textbf{Caus.} & \textbf{Nar.} & \textbf{Char.} & \textbf{Theme} & \textbf{Goal.} & \textbf{Soc.} & \textbf{Hypo.} \\
\midrule
VideoChat Flash (7B) & 27.22	&30.01&	32.21&	31.25&	30.79&	32.38&	28.35\\
Qwen2.5VL (7B) & 31.15&	34.97&	34.71&	35.56&	33.10&	33.88&	\best{33.22}\\
\midrule
\rowcolor{gray!15}
Video-NaRA & \best{34.92} & \best{37.87} & \best{36.91} & \best{38.68} & \best{33.98} & \best{35.68} & 32.11\\

\bottomrule
\end{tabular}
\end{table}


\section{Ablation Studies}

\subsection{Role of Narrative Memory Bank}
We conducted ablation to understand the contribution of memory bank and retriever module. As shown in \autoref{tab:performance}, the Full Model consistently achieves the best results across all distances—Short, Medium, and Far—demonstrating the importance of combining both narrative structure and retrieval. Removing Narrative Memory Bank leads to a drop across all four metrics. The decline is most visible in the evidence retrieval of Far setting (e.g., evidence drops (35.20 → 32.70)), indicating that narrative grouping is particularly crucial when evidence is temporally distant. Removing CLIP-based retrieval further reduces Evidence and Reasoning accuracy, showing that retrieval is critical for grounding answers in correct clips. Largest degradation occurs when both components are removed. Overall, these trends show that narrative structure provides a strong prior, helping the model maintain temporal coherence, while retrieval drives precise grounding. Their combination yields best reasoning performance, especially in long-range settings where event dependencies span large portions of video.

\subsection{Effect of scene distance}

We group questions by the distance between the question location and its supporting evidence (Short, Medium, Far) to study how temporal separation affects performance. In the long form video understanding, locating the exact portion of the video to answer the questions become critical. With the evidences in short, medium, and far, the  model has to locate evidences spread across the video. We can see from the \autoref{tab:performance} that locating even short distance evidence which are grouped together is difficult for the model. Interestingly, \autoref{tab:performance} does not show a clear degradation with increasing distance—in many cases, Medium and Far questions obtain scores comparable to or slightly higher than Short ones.

\subsection{Reasoning Capabilities with complexity}
NA-VQA tests advanced reasoning abilities by requiring models to connect events, understand motivations, and identify cause-and-effect relationships throughout long videos. We analyzed how different models perform various reasoning tasks~\autoref{tab:reasoning}. Video-NaRA achieves the strongest performance across nearly all categories, demonstrating more stable reasoning compared to baseline VLMs. Traditional VLMs such as VideoChat-Flash and Qwen2.5-VL show larger variance across categories—for example, performing relatively better on Character and Theme reasoning but struggling on Causal. Overall, the results indicate that NA-VQA exposes genuine reasoning challenges: models are not uniformly good across categories, and the hardest tasks remain those requiring multi-step inference (causal chains), narrative understanding (themes/goals), and counterfactual reasoning. Video-NaRA’s improvements highlight the effectiveness of organizing information into narrative structure before performing retrieval and reasoning.  


\section{Conclusion}

In this work, we introduced NA-VQA, a benchmark designed to evaluate the reasoning capabilities of MLLMs on full-length videos. Our experiments show that current models struggle to locate and integrate the correct evidence, highlighting that reasoning failures stem more from retrieval limitations than context length. To address this, we proposed Video-NaRA, a narrative-based construction and retrieval framework that consistently outperforms existing approaches. Overall, NA-VQA provides a rigorous foundation for studying narrative reasoning long videos. We hope this benchmark and framework will catalyze future research in very long-form complex video understanding, narrative modeling, and memory-augmented multimodal systems.

\small
\bibliographystyle{ieeenat_fullname}
\bibliography{main}


\end{document}